\documentclass[letterpaper, 10 pt, conference]{ieeeconf}  

\IEEEoverridecommandlockouts                              

\usepackage{graphicx} 
\usepackage{amsmath} 
\usepackage{amssymb}  
\usepackage{xcolor}
\usepackage{multirow}
\usepackage{makecell}
\usepackage{algorithm}
\usepackage{algpseudocode}
\usepackage[utf8]{inputenc}
\makeatletter
\let\NAT@parse\undefined
\makeatother
\usepackage{hyperref}

\title{\LARGE \bf
GLAMDRING: \underline{G}ait \underline{L}earning \underline{A}nd \underline{M}orphology co-\underline{D}esign via \underline{R}einforcement Learn\underline{ING} of CPGs
}

\author{Amogh Joshi and Kaushik Roy\\
Purdue University, West Lafayette, IN 47907, USA\\$\{joshi157, kaushik\}@purdue.edu$%
\vspace{-5mm}
}

\begin{document}

\maketitle
\thispagestyle{empty}
\pagestyle{empty}

\begin{abstract}

Robots are moving out of the structured factory floor and into unstructured environments such as disaster sites, planetary surfaces, and agricultural fields, for which the right robot often does not yet exist. We present GLAMDRING, a framework that synthesizes the optimal robot for a locomotion task and, jointly, learns the controller that drives it. For the given specifications of forward-velocity bounds, a per-actuator power budget, an actuator library, and a payload requirement, GLAMDRING returns a matched quadruped morphology (link geometry and per-joint actuators) and a Hopf-oscillator Central Pattern Generator (CPG) gait policy. We rank feasible designs against a target design objective, viz., maximum speed, minimum Cost of Transport (CoT), or max Payload Margin. Because body and locomotion are coupled, the optimal morphology dictates how a robot is driven, while optimal gait depends on the physical body. We train a small number of CPG policies by reinforcement learning across the space of candidate morphologies, co-learning the gait with the underlying robot hardware. Link lengths and actuators are then resolved post-hoc from the policy's logged operating envelope, reducing synthesis cost to a small, fixed number of reinforcement-learning runs instead of one per candidate. Our experiments show three key findings: co-designing body and gait is necessary to satisfy locomotion constraints; actuator-envelope feasibility, rather than locomotion success alone, determines realizable payload capacity; and canonical animal gaits emerge naturally in most designs from morphology and constraints alone. A real-world demonstration further highlights the efficacy of our work.

\end{abstract}

\section{INTRODUCTION} \label{sec:introduction}
For most of their history, robots were stationary pieces of equipment bolted to the floor, operating far from people, and repeatedly performing the same task in a controlled, highly structured environment. However, this is no longer the case for today's robots. Modern robots increasingly operate alongside humans on ill-defined tasks in unstructured environments hitherto restricted to humans. Warehouse robots pick, place, and sort items alongside workers, delivery robots cross public roads and walkways, surgical robots operate on patients {\em in vivo}, and autonomous cars share roads with human drivers, and the rate at which they are deployed continues to grow year after year \cite{ifr2023worldrobotics, yang2018grandchallenges}.

Conventional robot design pipelines treat morphology design and controller design as two sequential and independent problems. Human engineers first specify the robot's morphology, i.e., limb geometry, joint layout, and actuator selection, and then iteratively refine the design for a target task using simulation and a large body of accumulated intuition \cite{seok2013design, shin2022hound}. Only after the morphology is finalized is a controller designed to produce the desired locomotion behavior.
This separation overlooks the strong coupling between morphology and control. A robot's dynamics and actuation limits constrain the behavior its controller can realize, while the controller and desired locomotion behavior impose constraints on the robot's physical design \cite{dinev2022versatile, fadini2024codesigning}. Consequently, optimizing morphology or control in isolation may exclude solutions that can only emerge through the co-design of morphology and control.

This limitation is especially important when the appropriate morphology is unknown. While established platforms can guide design for well-defined tasks, robots are increasingly deployed in unstructured environments such as collapsed buildings, caves, and decommissioned industrial sites, which differ substantially from the environments for which existing platforms were developed \cite{yang2018grandchallenges}. In these settings, fixing morphology before controller design may impose arbitrary constraints on the solution. Therefore, the challenge is both to determine how the robot should act and how to build it.

This motivates a shift from sequential robot design toward \textit{robot synthesis}: the constrained joint optimization of morphology and control for a specified task. Rather than selecting a robot body {\em a priori} and subsequently designing a controller around it, proposed robot synthesis treats physical design and the control policy as coupled variables, searching for robot–controller pairs that best satisfy the task objective.

Legged morphologies are a natural target for \textit{robot synthesis}. Unlike wheeled or tracked systems, legs can exploit footholds, adapt body posture to terrain, and support diverse behavior such as walking, climbing, jumping, and crawling on a common mechanical platform \cite{raibert1986legged, hutter2016anymal}. Biological systems further illustrate the versatility of legged locomotion across challenging terrain and operating regimes. Quadrupedal animals can negotiate steep rock, transition between gaits across speeds, and maintain high-speed locomotion over uneven ground \cite{dickinson2000howanimals, full1999templates, alexander2003principles}. Consequently, legged design spaces offer both broad locomotor capability and biologically grounded priors for synthesis in unstructured settings.

Learning-based control offers a promising route to morphology–control co-design. While model-based methods typically require accurate dynamics models and manually specified gait structures \cite{dicarlo2018dynamic, winkler2018gait}, deep reinforcement learning can learn robust locomotion policies across challenging terrain and natural environments through interaction only, requiring less task-specific structure \cite{hwangbo2019agile, lee2020learning, miki2022learning, kumar2021rma}.
\begin{figure*}[ht]
    \centering
    \includegraphics[width=\textwidth]{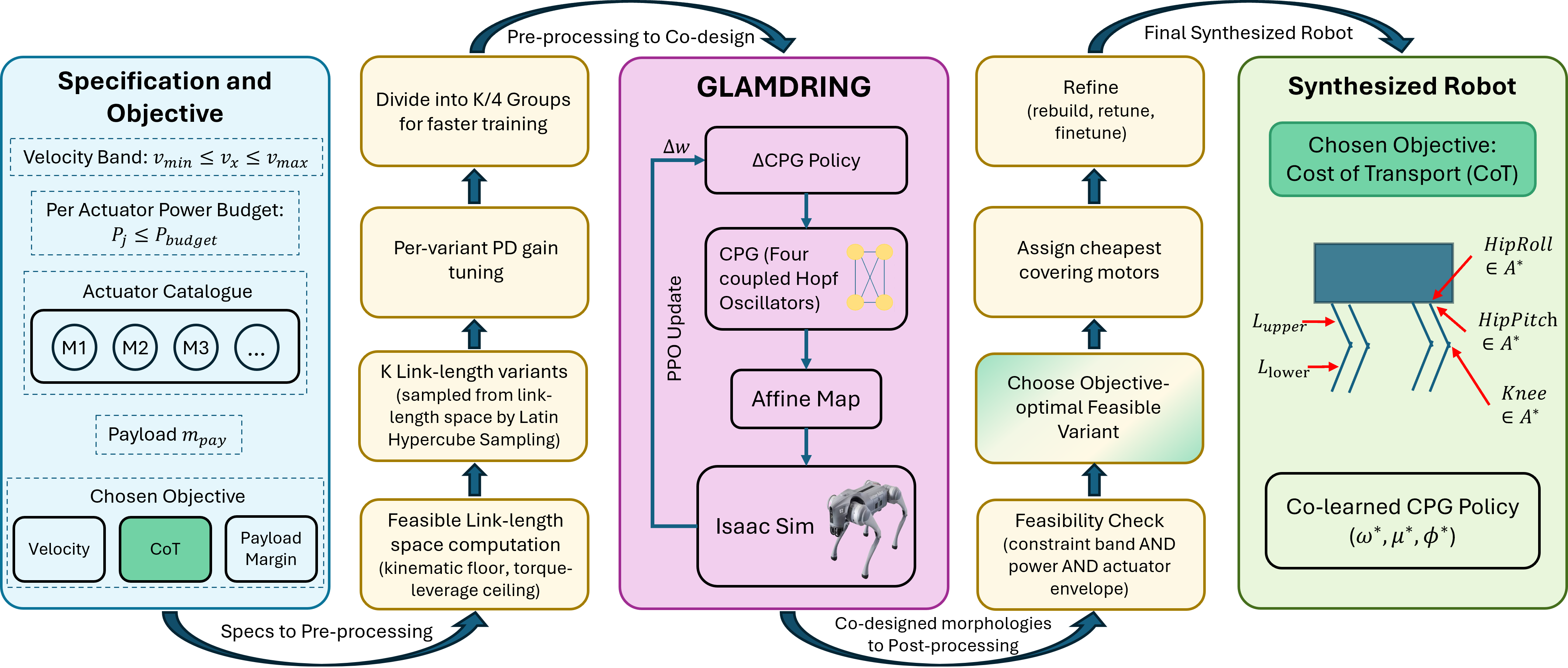}
    \caption{Overview of GLAMDRING. A user specification, i.e., a target velocity band, payload, per-actuator power budget, actuator catalog, and design objective, defines a constrained synthesis problem. GLAMDRING computes admissible link-length bounds, samples candidate morphologies, partitions them into groups of four to bound per-run GPU memory and simulation cost, and trains one morphology-conditioned CPG policy per group in simulation. Each policy emits incremental updates to Hopf-oscillator and affine joint-map parameters, while post-training rollouts log joint torque-speed operating points. Feasible designs are then filtered by velocity, power, and actuator-envelope constraints; the objective-optimal feasible morphology is selected, assigned the lowest-cost motors whose envelopes cover its operating points, and refined with the selected actuator masses. The output is a matched link geometry, actuator assignment, and CPG policy.}
    \label{fig:overview}
    \vspace{-5mm}
\end{figure*}
These results support extending optimization from control to mechanical design. Prior work has jointly learned simulated bodies and controllers through evolutionary search and reinforcement learning \cite{sims1994evolving, gupta2021embodied, ha2019reinforcement, schaff2019jointly, yuan2022transform2act}, but typically over virtual morphology spaces that do not account for component availability, actuation limits, power and payload budgets, or hardware feasibility, as detailed in Related Work.

Controller choice is central to morphology–control synthesis: it must generate effective locomotion across candidate morphologies while remaining compact enough for repeated evaluation. Central Pattern Generators (CPGs), bio-inspired oscillator networks that represent gait with a small set of interpretable parameters such as frequency, amplitude, and inter-limb phase offsets \cite{ijspeert2008cpg, ijspeert2007salamander, righetti2008pattern, owaki2017gait}, meet this need: the same controller structure can be evaluated across candidate bodies without separately tuning a gait for each, and its learned parameters remain an interpretable description of each synthesized body's gait (Sec.~\ref{subsec:cpg_background}).

This paper presents GLAMDRING (Gait Learning And Morphology co-Design via Reinforcement LearnING of CPGs), a framework for jointly synthesizing a quadruped morphology and its gait controller for straight-line walking. Given a target velocity range, a per-actuator power budget, a payload, an actuator library, and a design objective, GLAMDRING returns a feasible morphology-controller pair. A small number of CPG policies are trained across candidate morphologies, enabling joint evaluation of body geometry and gait without per-design retraining or manual gait design. Link lengths and actuators are resolved post-hoc from the policy's logged operating envelope, reducing synthesis cost to a small, fixed number of reinforcement-learning runs rather than one per candidate.

\noindent The primary contributions of this work are:
\begin{itemize}
    \item A framework (GLAMDRING) for joint synthesis of legged morphology and CPG-based gait control under hard velocity, power, payload, and actuator constraints.
    \item Experiments demonstrating that co-designing body and gait is necessary to satisfy locomotion constraints.
    \item An actuator-envelope analysis showing that locomotion success under payload does not guarantee actuator feasibility, making actuator modeling necessary for design certification.
    \item Canonical animal gaits emerge in a substantial fraction of synthesized designs from morphology and constraints alone, without gait-specific rewards.
\end{itemize}

\section{RELATED WORK} \label{sec:rel_work}
To the best of our knowledge, no prior work jointly synthesizes a physically realizable legged robot by optimizing morphology, actuator selection, and a learned rhythmic controller under hard locomotion constraints. GLAMDRING integrates these elements in a single morphology-gait co-design framework, returning a feasible robot-controller pair optimized for a specified objective. We position the method against three related research threads, as shown in Table \ref{tab:capability}.

\begin{table*}[ht]
\centering
\caption{Capability comparison: GLAMDRING is the only method with hard velocity, power, payload, and actuator-envelope constraints combined.}
\label{tab:capability}
\setlength{\tabcolsep}{3pt}
\begin{tabular}{|c|c|c|c|c|c|c|c|}
    \hline
    Method & $v$ floor (hard) & Power/Energy & Actuator model & Payload & Link geom. & Controller & Constraint handling \\
    \hline
    Ha et al.\ \cite{ha2018computational}              & $\sim$ & $\sim$ & --                   & --              & \checkmark & traj.\ opt.    & soft \\
    Vitruvio \cite{chadwick2020vitruvio}               & --     & \checkmark & limb + gear       & --              & \checkmark & fixed          & GA-selected \\
    Dinev et al.\ \cite{dinev2022versatile}            & --     & \checkmark & motor mass + gear & \checkmark      & \checkmark & traj.\ opt.\ (DDP) & hard (design) \\
    Fadini et al.\ \cite{fadini2021computational}      & --     & \checkmark & full electro-mech. & --              & scale + gear & traj.\ opt.\ (DDP) & soft / penalty \\
    KAIST HOUND \cite{shin2022hound}                   & \checkmark & \checkmark & gear train (MINLP) & $\sim$       & fixed      & NMPC           & hard \\
    CPG-RL \cite{bellegarda2022cpgrl}                  & reward & reward & -- (needs IK)    & tested (115\%)  & -- (fixed) & learned CPG    & soft \\
    \textbf{GLAMDRING (ours)}                           & \textbf{\checkmark} & \textbf{\checkmark} & \textbf{library + $\tau$--$\omega$ env.} & \textbf{\checkmark} & \textbf{\checkmark} & \textbf{learned CPG (no IK)} & \textbf{hard (post-hoc)} \\
    \hline
    \end{tabular}
\begin{flushleft}\footnotesize
    \checkmark\ = full, -- = none, $\sim$ = partial.
\end{flushleft}
\vspace{-6mm}
\end{table*}

\subsection{Model-based morphology-control co-design}
Several works jointly optimize parameterized robot designs and model-based motions for prescribed locomotion behaviors. Ha et al.~\cite{ha2018computational} optimize link lengths and actuator placements together with trajectories, actuator inputs, and contact forces. Dinev et al.~\cite{dinev2022versatile} differentiate through a constrained motion planner to optimize quadruped geometry, payload distribution, motor mass, and gear ratio for trotting and jumping. Fadini et al.~\cite{fadini2021computational,fadini2024codesigning} combine genetic search with trajectory optimization to optimize robot scale, actuation, and motion timing using detailed electro-mechanical loss models. Related work differentiates whole-body controllers with respect to design variables \cite{devincenti2021controlaware}, seeks robust designs across task scenarios \cite{bravopalacios2020onerobot,bravopalacios2022admm}, optimizes geometry and transmission with fixed control \cite{chadwick2020vitruvio}, or designs gear trains under discrete and thermal constraints \cite{shin2022hound}. These methods demonstrate the benefits of morphology-control co-design, but use trajectory optimization, MPC, or whole-body control rather than learned rhythmic control. They also generally optimize parameterized actuator properties or scaling factors, rather than selecting actuators from a discrete library while enforcing actuator operating envelopes, a velocity range, and payload constraints throughout synthesis.

\subsection{Learned CPG control with fixed morphology}
Learning-based CPG methods have demonstrated robust locomotion on fixed morphologies. Bellegarda and Ijspeert~\cite{bellegarda2022cpgrl} learn to modulate CPG amplitude and frequency on a Unitree A1, demonstrate sim-to-real transfer, and evaluate robustness to a 13.75~kg added load. Related work learns to modulate trajectory generators \cite{iscen2018pmtg} or optimizes CPG-based locomotion controllers for fixed morphologies \cite{ijspeert2007salamander}. In these works, morphology and actuator selection are fixed; oscillator outputs are also commonly mapped to joint targets through inverse kinematics. GLAMDRING instead jointly optimizes link geometry and actuator selection with a learned CPG controller, including a learnable affine oscillator-to-joint map that removes the inverse kinematics (IK) solver.

\subsection{Learning to co-design bodies and controllers}
A separate literature jointly optimizes simulated bodies and controllers through evolutionary search \cite{sims1994evolving,gupta2021embodied}, reinforcement learning over parameterized or editable morphologies \cite{ha2019reinforcement,schaff2019jointly,yuan2022transform2act}, terrain-driven topology search \cite{zhao2020robogrammar}, surrogate-based optimization \cite{chen2022c2}, and meta-RL adaptation across commands or terrains \cite{belmontebaeza2022metarl}. These approaches demonstrate joint body-controller optimization, but generally search virtual morphology spaces using reward-based objectives. They do not target quadruped synthesis from a physical actuator library subject to actuator torque-speed and power envelopes, payload requirements, and hardware feasibility constraints.

\section{GLAMDRING FRAMEWORK} \label{sec:framework}
GLAMDRING synthesizes a quadruped morphology and rhythmic gait controller for straight-line walking. The user specifies a target forward-velocity range, a payload requirement, a per-actuator power budget, an actuator library, and a \textit{design objective}, which is the criterion used to rank designs that satisfy all specified constraints (e.g., maximum mean velocity, minimum cost of transport, or maximum payload margin). GLAMDRING returns the feasible morphology-controller pair that optimizes this objective.

Rather than fixing morphology before controller design, GLAMDRING trains a morphology-conditioned CPG policy across candidate link geometries. The resulting gait behavior is evaluated jointly with morphology and actuator feasibility within a single synthesis procedure. The framework comprises a CPG-based gait representation, morphology-conditioned reinforcement learning, and actuator-envelope-based feasibility and selection, as shown in Algorithm \ref{alg:glamdring}.

\subsection{Problem Formulation} \label{subsec:problem}
We consider a quadruped with continuous upper- and lower-leg lengths and discrete actuator assignments:
\begin{equation}
\mathbf{l}=(l_{\mathrm{upper}},l_{\mathrm{lower}})\in\mathcal{L}, \quad
\mathbf{a}=(a_{\mathrm{hr}},a_{\mathrm{hp}},a_{\mathrm{k}})\in\mathcal{A}^{3},
\end{equation}
where $\mathcal{L}$ is the admissible link-geometry space and $\mathcal{A}$ is a library of physical actuators, both defined in Section~\ref{subsec:synthesis}. The gait controller is a CPG policy $\pi_{\theta}$.

\begin{algorithm*}[ht]
\caption{GLAMDRING Robot Synthesis}
\label{alg:glamdring}
\small
\begin{algorithmic}[1]
\Require Spec $S = \{v_{\min},\, v_{\max},\, P_{\mathrm{max}},\, \text{catalogue } \mathcal{A},\, \text{payload } m_p\}$;
\Statex \quad objective $\mathrm{obj} \in \{\text{speed, CoT, payload-margin}\}$; number of variants K (default: 32); group size $g_s$ (default: 4)
\Ensure One design $(\mathbf{l}^{*},\, \mathbf{a}^{*},\, \pi^{*})$, or \textsc{infeasible}
\Procedure{Synthesize}{$S,\, \mathrm{obj},\, K$}
    \State $(\mathbf{l}_{\mathrm{lo}},\, \mathbf{l}_{\mathrm{hi}}) \gets \textsc{LinkBounds}(\mathcal{A},\, S)$ \Comment{kinematic clearance, torque-leverage vs.\ strongest motor in $\mathcal{A}$}
    \State $V \gets \textsc{LatinHypercube}(\mathbf{l}_{\mathrm{lo}},\, \mathbf{l}_{\mathrm{hi}},\, K)$ \Comment{$K$ link-length variants}
    \State \textbf{for} $v \in V$: \; $G[v] \gets \textsc{TunePD}(v,\, m_p)$ \Comment{per-variant PD gains}
    \State $\{V_1, \ldots, V_{K/g_s}\} \gets \textsc{Partition}(V,\, g_s)$ \Comment{caps concurrent geometries: GPU memory \& sim speed}
    \State \textbf{for} $g = 1 \ldots K/g_s$: \Comment{one training run per group}
    \State \quad $\pi_g \gets \textsc{InitPolicy}()$;\; $\mathrm{CPG} \gets \textsc{HopfNet}(N,\, \phi \gets \mathrm{trot})$
    \State \quad \textbf{for} $\mathrm{iteration} = 1 \ldots T_{\mathrm{ppo}}$: \Comment{PPO; envs round-robin over $V_g$ (domain randomization)}
    \State \quad\quad \textbf{for} $t = 1 \ldots H$:
    \State \quad\qquad $\Delta \gets \pi_g(s_t)$ \Comment{corrections to $(\omega,\, \mu,\, \phi,\, s_{\square},\, b_{\square})$}
    \State \quad\qquad $(\omega,\, \mu,\, \phi,\, s_{\square},\, b_{\square}) \gets \mathrm{clamp}(\mathrm{prev} + \Delta)$
    \State \quad\qquad $(x,\, y) \gets \mathrm{CPG.step}(\omega,\, \mu,\, \phi,\, k)$ \Comment{Eq.~\ref{eq:coupled_hopf}, forward Euler at 100\,Hz}
    \State \quad\qquad $\theta_{\mathrm{target}} \gets \textsc{AffineMap}(x,\, y,\, s_{\square},\, b_{\square})$ \Comment{Eq.~\ref{eq:joint_setpt_map} - no inverse kinematics}
    \State \quad\qquad $s_{t+1},\, r_t \gets \textsc{Env.step}(\theta_{\mathrm{target}})$ \Comment{$r_t$ from Eq.~\ref{eq:reward}; ramped constraint penalties}
    \State \quad $\pi_g \gets \textsc{PPO.update}()$
    \State $E \gets \bigcup_g \textsc{LogEnvelope}(\pi_g,\, V_g)$ \Comment{per-joint-type $(|\dot{q}|,\, \max|\tau|)$ bins over all $K$ variants}
    \State $F \gets \{\, v \in V : \textsc{Feasible}(v;\, S)\,\}$ \Comment{band $\wedge$ envelope $\wedge$ $P_{\mathrm{max}}$}
    \State \textbf{if} $F = \emptyset$: \; \Return \textsc{infeasible}
    \State $\mathbf{l}^{*} \gets \arg\max_{v \in F}\, \textsc{Objective}(v;\, \mathrm{obj})$ \Comment{selects the feasible variant maximizing the specified objective}
    \State $\mathbf{a}^{*} \gets \bigcup_{j \in \{\mathrm{hr,\, hp,\, k}\}} \arg\min_{a \in \mathcal{A}}\, \mathrm{cost}(a)$ \;\textbf{s.t.}\; $\textsc{Covers}(a,\, E[j],\, \mathrm{margin})$
    \State \textbf{if} $\mathbf{a}^{*}$ incomplete: \; \Return \textsc{infeasible} \Comment{learned gait exceeds every catalogue motor for joint $j$}
    \State $(\pi^{*},\, \mathbf{a}^{*}) \gets \textsc{Refine}(\mathbf{l}^{*},\, \mathbf{a}^{*},\, m_p)$ \Comment{rebuild with true masses, re-tune PD, warm-start from $\pi_g \ni \mathbf{l}^{*}$, re-select}
    \State \Return $(\mathbf{l}^{*},\, \mathbf{a}^{*},\, \pi^{*})$
\EndProcedure
\end{algorithmic}
\end{algorithm*}

The synthesis specification includes a target velocity interval
$\mathcal{V}=[v_{\min},v_{\max}]$, a required payload $m_{\mathrm{p}}$, a per-actuator power limit $P_{\max}$, and a scalar design objective $J$. The payload is incorporated into the simulation during training and evaluation; $m_{\mathrm{p}}=0$ when no payload is specified. GLAMDRING seeks to satisfy:
\begin{equation}
\begin{aligned}
(\mathbf{l}^{*},\mathbf{a}^{*},\theta^{*})
= \arg\max_{\mathbf{l},\mathbf{a},\theta}\quad
& J(\mathbf{l},\mathbf{a},\pi_{\theta})\\
\mathrm{s.t.}\quad
& \bar{v}_x(\mathbf{l},\mathbf{a},\pi_{\theta};m_{\mathrm{p}})
\in [v_{\min},v_{\max}],\\
& P_j(\mathbf{l},\mathbf{a},\pi_{\theta};m_{\mathrm{p}})
\leq P_{\max}, \quad \forall j,\\
& (|\tau_j|,|\dot{q}_j|)
\in \mathcal{E}(a_j), \quad \forall j,\\
& \mathbf{l}\in\mathcal{L}, \qquad
\mathbf{a}\in\mathcal{A}^{3}.
\end{aligned}
\label{eq:synthesis_problem}
\end{equation}
where, $\bar{v}_x$ is the mean forward velocity, $P_j$ is the power required at the joint $j$, and $\mathcal{E}(a_j)$ denotes the torque-speed envelope of the actuator $a_j$. The objective $J$ ranks only combinations that satisfy all hardware and locomotion constraints. For the minimum-CoT objective, we use the dimensionless cost of transport $CoT=\frac{P}{m_{tot}g\bar{v}_x}$, where $P$ is the time-averaged total actuator power over the evaluation rollout and $m_{tot}$ is the total carried mass of the robot and specified payload. CoT measures the energy required to move unit weight over unit distance, thereby normalizing energy use across candidate morphologies, actuator assignments, payloads, and operating speeds. We evaluate and compare CoT only for variants that satisfy the target velocity band and remain upright; otherwise, low or undefined forward progress can make the ratio uninformative.

\begin{table}[hb]
\centering
\caption{CPG parameters. Fixed values are set at initialization; learned parameters are updated using policy outputs.}
\label{tab:cpg_params}
\begin{tabular}{|c|c|c|c|}
\hline
    Symbol & Parameter & Learned? & Range/Value \\
    \hline
    $\alpha$ & \makecell{Hopf convergence\\rate} & no  & 10 \\
    $k$ & Coupling gain & no  & 1 \\
    $\omega$ & \makecell{Angular frequency\\(rad/s)} & yes & \makecell{$[1.0,\,15.0]$,\\init.\ 5.0} \\
    $\mu$ & Squared amplitude & yes & \makecell{$[0.1,\,4.0]$,\\init.\ 1.0} \\
    $\phi_{ij}$ & \makecell{Inter-leg\\phase offset}          & yes & \makecell{$(-\pi,\,\pi]$,\\ skew-sym., 6 free} \\
    $s_{\mathrm{hr}}, s_{\mathrm{hp}}, s_{\mathrm{k}}$ & \makecell{Joint-map scale\\(rad/unit)} & yes & $[-1.5,\,1.5]$ \\
    $b_{\mathrm{hr}}$ & \makecell{Joint-map offset,\\hip roll} & yes & \makecell{joint limits,\\init.\ 0.0} \\
    $b_{\mathrm{hp}}$ & \makecell{Joint-map offset,\\hip pitch} & yes & \makecell{joint limits,\\init.\ 0.0} \\
    $b_{\mathrm{k}}$        & \makecell{Joint-map offset,\\knee}          & yes & \makecell{joint limits,\\init. knee angle} \\
    \hline
\end{tabular}
\vspace{-5mm}
\end{table}


\subsection{Central Pattern Generators} \label{subsec:cpg_background}
Central pattern generators (CPGs) are networks of coupled oscillators that generate coordinated rhythmic activity. In biological locomotion, CPGs are spinal circuits that produce the basic locomotor rhythm, while descending commands and sensory feedback modulate frequency, amplitude, and inter-limb coordination \cite{ijspeert2008cpg,ijspeert2007salamander}. Robotic CPGs provide an analogous low-dimensional representation, with gait characterized by interpretable parameters such as frequency, amplitude, and phase offsets \cite{righetti2008pattern,owaki2017gait}. A more detailed list of CPG parameters, along with their standard values, is shown in Table \ref{tab:cpg_params}. Their parameters can be continuously modulated to express different stride magnitudes, frequencies, and coordination patterns, making them well suited for co-design, as a single CPG controller structure can be evaluated across candidate bodies without separately authoring a gait for each.

\subsection{GLAMDRING CPG Controller} \label{subsec:cpg_controller}

GLAMDRING uses four coupled Hopf oscillators, one per leg. Oscillator $i$ has Cartesian state $(x_i,y_i)$, which avoids the $r{=}0$ singularity of polar coordinates and provides a differentiable linear read-out to joint targets. Each oscillator evolves according to
\begin{equation}
    \begin{aligned}
            \dot{x}_i = \alpha(\mu - r_i^2)x_i - \omega y_i + \sum_{j \neq i} k [ x_j \cos \phi_{ij} - y_j \sin \phi_{ij} ]
            \\
            \dot{y}_i = \alpha(\mu - r_i^2)y_i + \omega x_i + \sum_{j \neq i} k [ x_j \sin \phi_{ij} + y_j \cos \phi_{ij} ]
    \end{aligned}
    \label{eq:coupled_hopf}
\end{equation}
where $r_i^2 = x_i^2 + y_i^2$. At steady state, each oscillator traces a circle of radius $\sqrt{\mu}$ at angular frequency $\omega$. The coupling terms entrain the network so that oscillator $j$ leads oscillator $i$ by the phase offset $\phi_{ij}$. We fix the convergence rate $\alpha$ and coupling gain $k$. The policy modulates $\omega$, $\mu$, and the six independent entries of the skew-symmetric phase matrix $\phi$.

Rather than mapping oscillator states to task-space trajectories and solving morphology-specific inverse kinematics \cite{bellegarda2022cpgrl}, GLAMDRING maps oscillator states directly to joint targets:
\begin{equation}
    \begin{aligned}
        \theta_{\mathrm{hr},i} &= s_{\mathrm{hr}}y_i+b_{\mathrm{hr}},\\
        \theta_{\mathrm{hp},i} &= s_{\mathrm{hp}}x_i+b_{\mathrm{hp}},\\
        \theta_{\mathrm{k},i} &= s_{\mathrm{k}}x_i+b_{\mathrm{k}}.
    \end{aligned}
    \label{eq:joint_setpt_map}
\end{equation}

\begin{figure}[t]
    \centering
    \includegraphics[width=\columnwidth]{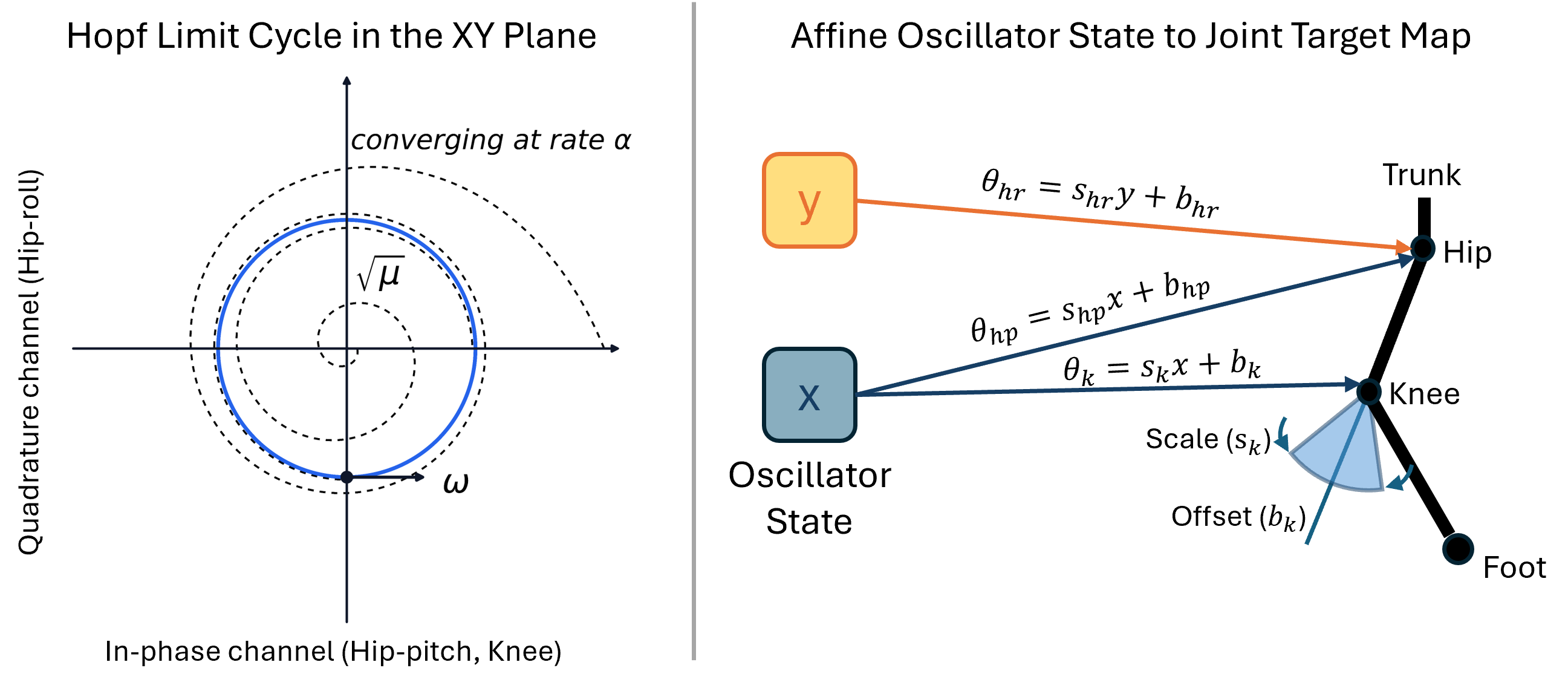}
    \caption{Left: Hopf oscillator limit cycle in the $(x,y)$ plane. The in-phase channel $x$ drives hip-pitch and knee; the quadrature channel $y$ drives hip-roll. Right: the learned affine map from oscillator state to one leg's joint targets. Scale and offset parameters $(s_{\square}, b_{\square})$ are policy outputs, eliminating the need for an inverse kinematics solver.}
    \label{fig:oscillator_map}
    \vspace{-5mm}
\end{figure}

Hip-roll uses the quadrature state $y_i$, whereas hip-pitch and knee use $x_i$. The scale and offset coefficients are shared across legs and learned jointly with the oscillator parameters. This affine joint-space readout avoids re-deriving an inverse-kinematics model for each candidate geometry. Figure~\ref{fig:oscillator_map} illustrates the oscillator limit cycle and the affine readout from oscillator state to one leg's three joints.

\subsection{Morphology-conditioned CPG Learning} \label{subsec:learning}
We train the CPG controller using PPO \cite{schulman2017ppo}. At each control step, the policy observes oscillator state, joint positions and velocities, base linear and angular velocity, projected gravity, the previous action, and normalized link lengths. It outputs bounded increments to the oscillator and joint-map parameters:
\begin{equation}
    \begin{aligned}
        a_t = (\Delta\omega,\Delta\mu,\Delta\phi_{\mathrm{upper}},
        \Delta s_{\mathrm{hr}},\Delta b_{\mathrm{hr}},
        \Delta s_{\mathrm{hp}},\Delta b_{\mathrm{hp}},
        \Delta s_{\mathrm{k}},\Delta b_{\mathrm{k}}).
    \end{aligned}
    \label{eq:action_out}
\end{equation}

Incremental actions preserve the stable rhythmic prior while allowing the policy to adapt frequency, amplitude, phase coordination, and joint motion to each geometry.

The reward promotes forward motion in the target velocity interval while penalizing power consumption, joint effort, falls, and abrupt action changes:
\begin{equation}
    \begin{aligned}
        r_t = r_{\mathrm{vel}}
        -w_P r_{\mathrm{power}}
        -w_{\tau}r_{\mathrm{effort}}
        -w_{\mathrm{smooth}}r_{\mathrm{smooth}}\\
        -r_{\mathrm{fall}}
        -\lambda_v(t)c_v
        -\lambda_P(t)c_P.
    \end{aligned}
    \label{eq:reward}
\end{equation}
The terms $c_v$ and $c_P$ penalize velocity and power requirement violations. Their weights ramp during training to avoid dominating exploration early. These penalties guide learning but do not replace final hardware-feasibility verification.

\subsection{Design Space and Feasibility} \label{subsec:synthesis}
\begin{itemize}
    \item \textbf{Admissible link geometry ($\mathcal{L}$).}
    The admissible geometry space is a rectangular box in $(l_{\mathrm{upper}}, l_{\mathrm{lower}})$, computed from geometry and statics alone. The lower bound requires that the foot at the deepest commanded knee flexion clears the hip gimbal by at least a fixed margin. The upper bound requires that the torque to hold the robot's weight at the worst-case (horizontal-leg) moment arm, plus a dynamic reserve for leg-swing acceleration, does not exceed the stall torque of the strongest actuator in the library; this bound is further capped at $0.75\, L_{\mathrm{body}}$. For the quadruped in this work, $\mathcal{L} = [0.118,\,0.540] \times [0.118,\,0.540]$~m. Candidate geometries are Latin-hypercube \cite{mckay1979lhs} sampled within $\mathcal{L}$; geometries that admit no stance-centering solution or exceed a 1.27:1 segment aspect ratio are resampled. Unless otherwise stated, GLAMDRING samples $K=32$ link-length variants per synthesis, balancing coverage of the morphology space against training runtime: smaller $K$ gives a coarser variant search, while larger $K$ increases coverage at additional computational cost. Because IsaacSim renders a distinct collision mesh per unique link geometry, training all $K$ variants concurrently exceeds practical GPU memory and slows simulation; GLAMDRING instead partitions variants into groups of four and trains one shared CPG policy per group, giving $K/4$ training runs per synthesis (8 by default) rather than a single run over all $K$ variants or $K$ separate per-variant runs.
    \item \textbf{Actuator library ($\mathcal{A}$).}
    The library comprises 11 actuators spanning two families: quasi-direct-drive BLDC modules (peak joint torques 8-120~Nm, including commercial units from T-Motor, Unitree, Xiaomi, and RobStride) and brushed DC servos (3.5-12~Nm). Each motor is characterized by a torque-speed envelope derived from its electrical parameters, viz., phase resistance, torque constant, bus voltage, gear ratio, and efficiency, and current and torque limits, which yields the maximum available joint torque at any joint speed.
    \item \textbf{Feasibility and selection.}
    A candidate is feasible if it satisfies the target forward-velocity band and per-actuator power limit, and if each joint type can be assigned an actuator from $\mathcal{A}$ whose torque-speed envelope covers all logged operating points with a 15\% safety margin. Unless explicitly noted otherwise, reported design statistics are computed only over this fully feasible subset. Among feasible combinations, GLAMDRING selects the design maximizing $J$ and assigns the lowest-cost qualifying actuator to each joint type. The selected robot is rebuilt with the resulting actuator masses and the policy is warm-start fine-tuned before final evaluation. If no candidate admits a feasible actuator assignment, the specification is reported infeasible for the available library.
\end{itemize}

\section{EXPERIMENTS} \label{sec:experiments}
\begin{table*}[ht]
\centering
\caption{Ablation: co-design vs.\ fixed Go2 geometry across three specifications. All values are the mean $\pm$ std over five seeds.}
\label{tab:ablation}
\begin{tabular}{|c|c|c|c|c|c|c|}
\hline
    \multirow{2}{*}{Spec} & \multicolumn{2}{c|}{GLAMDRING (co-design)} & \multicolumn{2}{c|}{Go2 + own motor} & \multicolumn{2}{c|}{Go2 + GLAMDRING's motor} \\
    \cline{2-7}
     & $v$ (m/s) & CoT & $v$ (m/s) & CoT & $v$ (m/s) & CoT \\
    \hline
    default       & $0.729 \pm 0.044$ & $0.398 \pm 0.016$ & $1.751 \pm 0.207$ &   NA$^\ddagger$   & $0.290 \pm 0.041$ & $1.344 \pm 0.032$ \\
    +50\% payload & $0.858 \pm 0.027$ & $0.238 \pm 0.009$ & $0.693 \pm 0.027$    & $0.269 \pm 0.036$ & $0.676 \pm 0.049$ & $0.289 \pm 0.025$ \\
    40~W          & $4.32 \pm 2.74$   & $0.172 \pm 0.079$ &     $0.028 \pm 0.009$     &   NA$^\dagger$   & $0.050 \pm 0.024$ & NA$^\dagger$ \\
    \hline
\end{tabular}
\begin{flushleft}\footnotesize
``$\dagger$'' denotes that the policy fails the velocity constraint; ``$\ddagger$'' denotes that the policy fails the power constraint, both implying an unusable CoT number.
\end{flushleft}
\vspace{-5mm}
\end{table*}

All experiments use Isaac Lab with the PhysX engine \cite{makoviychuk2021isaacgym,mittal2023orbit}, using the default $K=32$ link-length variants in groups of four ($\S$\ref{subsec:synthesis}, Algorithm \ref{alg:glamdring}), and a minimum-velocity constraint of $v_{\min}=0.3$~m/s unless otherwise stated. Unless noted, reported statistics are computed over the subset of sampled variants satisfying all feasibility constraints, i.e., the target forward-velocity band, the per-actuator power limit, and the selected actuators' torque-speed envelopes; reported infeasibility rates and feasibility counts (fraction/number of sampled designs meeting the feasibility criteria of $\S$\ref{subsec:synthesis}), and explicitly labeled pre-filter results, instead use the full sampled population. The robot is a quadruped with a Go2-proportioned trunk ($0.72 \times 0.18 \times 0.22$~m, $\approx$7~kg) and four legs, each with a 2-DoF hip and a knee, drawing actuators from the library defined in $\S$.~\ref{subsec:synthesis} (11 units: 9 quasi-direct-drive BLDC modules spanning 8-120~Nm, 2 brushed DC servos spanning 3.5-12~Nm).

We report CoT as defined in $\S$ \ref{subsec:problem} and evaluate it only for variants that remain upright and satisfy the target velocity. As shown in ($\S$\ref{sec:rel_work}), a fair head-to-head benchmark against prior co-design methods does not exist. The comparisons below instead either fix geometry to a real platform ($\S$\ref{subsec:ablation}) or reproduce a related evaluation protocol ($\S$\ref{subsec:payload}) against our method under matched conditions. Unless otherwise stated, GLAMDRING selects the minimum-CoT design from the fully feasible subset.

\subsection{Co-Design Is Necessary} \label{subsec:ablation}
We test whether link length and actuators can be chosen separately by fixing the geometry to a real robot's proportions and training only the controller. Link length is set to a scaled Unitree Go2's leg dimensions: 0.213~m thigh and calf in the real URDF, scaled to $\approx$0.219~m to match our trunk's aspect ratio. Two fixed actuators are tested: the Go2's own motor (GO-M8010-6) and the actuator that GLAMDRING's co-design converges to. Only the CPG controller is trained; geometry is fixed to the real Go2, not GLAMDRING's search. This comparison is run at three specifications: the default setting, 50\% payload (9.83~kg added), and a 40~W per-actuator power budget.

Table~\ref{tab:ablation} reports GLAMDRING's co-designed solution against the fixed-geometry baselines; failed baselines report velocity only, indicating the failure mode, with CoT omitted. Under the strict 40~W budget, neither Go2 geometry satisfies the velocity constraint, making GLAMDRING the only feasible solution; its large velocity variation indicates that seeds discover substantially different low-power behaviors, yet the selected policies remain efficient. In the default setting, high-power policies often violate the torque-speed envelope and are marked infeasible, so GLAMDRING's feasible design is drawn from the remaining lower-velocity variants; the Go2 geometry with its own motor also reaches high speed but violates power, while with GLAMDRING's motor it is feasible but slow and inefficient. Under 50\% payload, all three methods are feasible, but co-design is again best. CoT need not increase monotonically with constraint tightness: payload directly enlarges $m_{\mathrm{tot}}$ in the CoT denominator ($\S$\ref{subsec:problem}), mechanically lowering CoT for comparable power and speed, independent of any absolute efficiency change. Also, each specification is solved via an independent, finite sample of $K$ morphologies ($\S$\ref{subsec:synthesis}) rather than an exhaustive search, so the 40~W minimum need not bound the default result; it simply reflects whichever low-power candidates that run's sample and training discovered. Overall, fixing geometry and swapping actuators can succeed in mild settings, but under tight constraints the feasible set depends jointly on morphology, actuation, and control.

\begin{figure}[hb]
\vspace{-4mm}
\includegraphics[width=\linewidth]{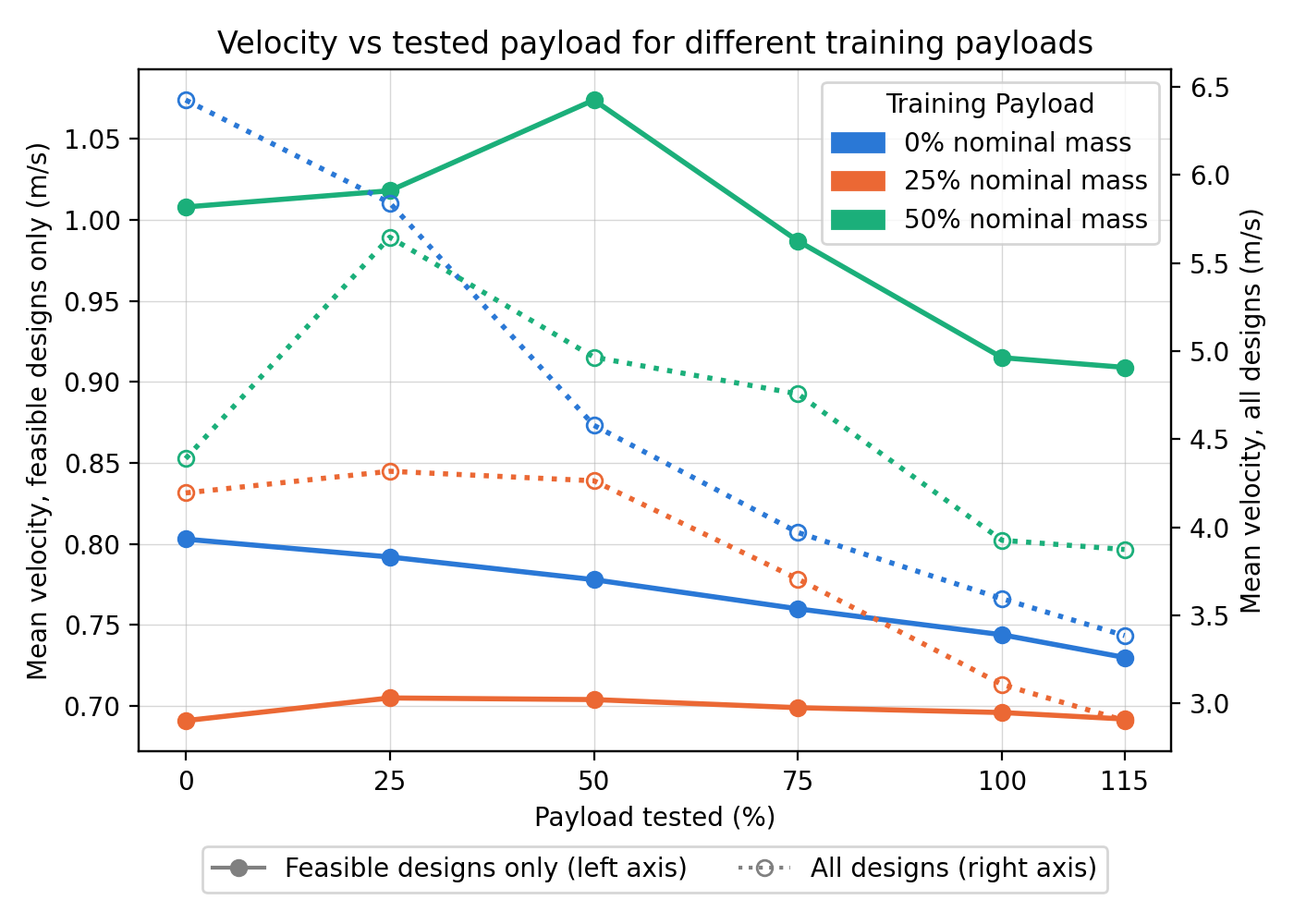}
\caption{Payload robustness with and without actuator-envelope filtering. For policies trained with 0, 25, and 50\% added payload, dotted curves (right axis) show mean velocity over variants satisfying the velocity and power checks, while solid curves (left axis) show the subset that also satisfies actuator torque-speed envelopes. The large gap between the two sets shows that high-velocity payload-robust gaits are often not actuator-feasible.}
\label{fig:payload}
\vspace{-5mm}
\end{figure}

\subsection{Velocity-Band Compliance Does Not Imply Actuator Feasibility} \label{subsec:payload}
We reproduce the CPG-RL payload test \cite{bellegarda2022cpgrl}: policies trained with added payloads of 0, 25, and 50\% of nominal mass (0, 4.92, 9.83~kg) are evaluated zero-shot with payloads of 0, 25, 50, 75, 100, and 115\% of nominal mass (0, 4.92, 9.83, 14.75, 19.67, 22.62~kg). For each design, we check velocity-band satisfaction, torque-speed envelope containment, and per-actuator power. Figure~\ref{fig:payload} compares mean forward velocity under each tested payload before and after actuator-envelope filtering. The dotted curves show that many learned CPG limit cycles retain locomotion-level performance under added payload, reproducing the robustness reported by CPG-RL \cite{bellegarda2022cpgrl}; the solid curves show the much smaller actuator-feasible subset.

This gap between viable gait and actuators widens with payload: in the 115\% payload test, only 3/32 zero-payload-trained variants remain actuator-feasible. This is the empirical case for actuator modeling in synthesis. CPG-RL identifies sim-to-real issues from ``unmodeled dynamics such as the actuators'' and ``lack of motor modeling'' \cite{bellegarda2022cpgrl}; GLAMDRING addresses that gap with an actuator-envelope check.

This separation also highlights objective-dependent selection: every GLAMDRING output is actuator-feasible for its specified payload. Under a zero-payload specification, speed- or CoT-optimized objectives may choose designs with little actuator margin that fail the out-of-spec 115\% payload check, whereas a payload-margin objective selects a certified design with enough reserve to remain buildable. Actuator modeling is what exposes this distinction.
\vspace{-0mm}

\subsection{Gait-Like Coordination Emerges Without Gait Rewards} \label{subsec:gaits}
GLAMDRING is never rewarded for any particular gait, yet the converged inter-leg phase offsets resemble recognized animal gaits in a significant number of runs. For every design we record the time-averaged phase matrix $\phi$, reduce it to the leading row of relative phases $\phi_{0j}/2\pi \bmod 1$, and classify by nearest neighbour against four canonical gaits: walk $[0, \tfrac{1}{2}, \tfrac{3}{4}, \tfrac{1}{4}]$, trot $[0, \tfrac{1}{2}, \tfrac{1}{2}, 0]$, pace $[0, \tfrac{1}{2}, 0, \tfrac{1}{2}]$, and bound $[0, 0, \tfrac{1}{2}, \tfrac{1}{2}]$. A random phase matrix sits about 0.25 from the nearest canonical gait, serving as the baseline.

Among reported runs, the strength of gait-like convergence depends strongly on the design specification driving the search, more than on gait type or speed. An illustrative example: two runs with identical link lengths, $(0.538,0.448)$~m, converge to different canonical phase patterns under different velocity constraints (Table~\ref{tab:gait_examples}). Across all reported feasible runs for the gait emergence experiments ($n=52$), inter-leg phase offsets ranged from near-perfect canonical matches ($\phi \hspace{1mm}dist=0.003$) to identifiable but non-canonical variants ($\phi \hspace{1mm}dist=0.17$), all meaningfully structured relative to the random baseline ($\phi \hspace{1mm}dist=0.25$)
We therefore interpret gait-like coordination as an emergent but condition-dependent property of the learned CPG design space. A full mechanistic account is left for future work.

\begin{figure}[b]
\centering
\vspace{-2mm}
\includegraphics[width=0.7\columnwidth]{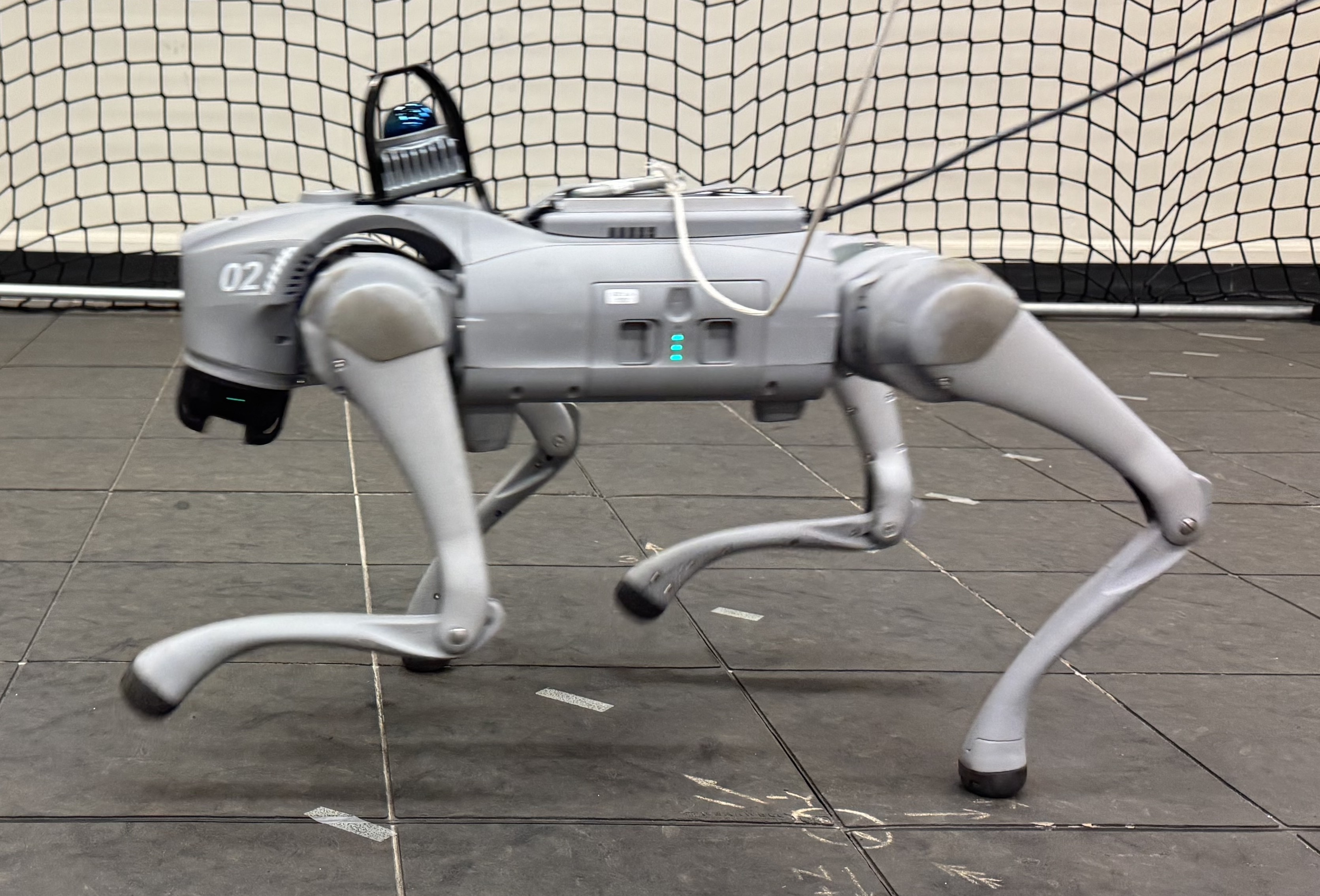}
\caption{Hardware deployment of the GLAMDRING-synthesised zero-payload CPG policy on an unmodified Unitree Go2. The robot walks stably in the Trot gait without retuning the learned gait parameters.}
\label{fig:go2_deploy}
\vspace{-6mm}
\end{figure}


\begin{table}[t]
\centering
\caption{Best match converged phase patterns per canonical gait. $(l_u,l_l)$ in metres; $\bar v$ in m/s; distance is $\phi_{0j}/2\pi \bmod 1$ nearest-neighbour distance.}
\label{tab:gait_examples}
\setlength{\tabcolsep}{4pt}
\begin{tabular}{|c|c|c|c|c|}
\hline
Gait & $(l_u,l_l)$ (m) & $\bar v$ & Dist. & $\phi_{0j}$ (rad) \\
\hline
Walk B & (0.468, 0.434) & 1.116 & 0.0504 & $[0,\,-3.10,\,-1.51,\,1.88]$ \\
Walk W  & (0.524, 0.537) & 0.315 & 0.1470 & $[0,\,-2.53,\,-1.51,\,0.88]$ \\
\hline
Trot B & (0.373, 0.303) & 0.555 & 0.0040 & $[0,\,-3.14,\,-3.14,\,0.03]$ \\
Trot W  & (0.417, 0.451) & 0.686 & 0.1789 & $[0,\,-2.80,\,-2.69,\,0.97]$ \\
\hline
Pace$^\dagger$ B & (0.538, 0.448) & 0.823 & 0.0145 & $[0,\,-3.14,\,0.09,\,3.13]$ \\
Pace W  & (0.531, 0.529) & 0.613 & 0.1095 & $[0,\,-2.73,\,-0.53,\,3.01]$ \\
\hline
Bound$^\dagger$ B & (0.538, 0.448) & 1.186 & 0.0026 & $[0,\,0.01,\,-3.14,\,3.13]$ \\
Bound W  & (0.427, 0.506) & 0.724 & 0.1251 & $[0,\,-0.57,\,-3.14,\,2.61]$ \\
\hline
\end{tabular}
\begin{flushleft}\footnotesize
$\dagger$Pace and Bound share link lengths but converge to different gaits under different design specifications, indicating that link geometry alone does not fix the nearest gait. Random baseline distance $\approx 0.25$.
\end{flushleft}
\vspace{-8mm}
\end{table}

\subsection{Hardware Deployment on Unitree Go2} \label{subsec:go2_deploy}
As a hardware validation, we instantiate GLAMDRING under a Go2-compatible specification: the design specifications are chosen such that the optimal solution converges to approximately the Unitree Go2's geometry and motors. This target is deliberately chosen, as fabricating a custom robot for each synthesised design is impractical, so we constrain the search to a geometry that is already commercially available and can be tested on unmodified hardware. The resulting zero-payload CPG policy is deployed on an unmodified Go2 without changing the learned gait parameters. The robot walks stably in hardware, as shown in Fig.~\ref{fig:go2_deploy} and the supplementary video.

This experiment is not intended as a full sim-to-real robustness study; rather, it verifies that the synthesised CPG controller and actuator-feasible assumptions produce a policy executable on a real commercial quadruped. It complements the fixed-geometry ablation (Section~\ref{subsec:ablation}), which shows that Go2-like geometry alone is not sufficient under all constraints: when GLAMDRING is given a Go2-compatible specification, the learned policy remains executable on the physical platform, whereas fixing the geometry without co-design can leave the controller outside the feasible set.

\section{CONCLUSION} \label{sec:conclusion}
We present GLAMDRING, a framework for joint synthesis of a quadruped morphology and its Central Pattern Generator (CPG) gait controller for straight-line locomotion. Given forward-velocity bounds, actuator power/payload constraints, and an actuator library, GLAMDRING returns a feasible morphology–controller pair optimized for speed, CoT, or payload margin. A network of Hopf-oscillator CPGs is trained via RL across candidate morphologies, while link geometry and actuator selection are resolved from the policy's logged torque–speed envelope using a small number of training runs, making synthesis more efficient than retraining per candidate design.

Experiments show that morphology and gait must be co-designed to satisfy the specified constraints, that learned gait robustness does not necessarily imply actuator robustness and therefore requires actuator-level certification, and that recognizable animal gaits emerge from the interaction of morphology, control, and physical constraints without gait-specific rewards. Together, these results demonstrate that constrained morphology–control co-design can produce physically realizable quadrupeds while exposing how body and gait jointly determine locomotion behavior.


\bibliographystyle{IEEEtran}
\bibliography{root}

\end{document}